\documentclass{edm_article}
\usepackage{hyperref}
\hypersetup{
	colorlinks=true,
	linkcolor=blue,
	filecolor=blue,      
	urlcolor=blue,
	citecolor=blue,
}
\usepackage{multirow} 
\begin{document}

\title{Using Vision Language Models to Detect Students' Academic Emotion through Facial Expressions}

\numberofauthors{1}
\author{
Deliang Wang, Chao Yang, Gaowei Chen\\
       Faculty of Education, The University of Hong Kong, Hong Kong\\
       \email{wdeliang@connect.hku.hk, chaoyang@connect.hku.hk, gwchen@hku.hk}
}

\maketitle

\begin{abstract}
Students' academic emotions significantly influence their social behavior and learning performance. Traditional approaches to automatically and accurately analyze these emotions have predominantly relied on supervised machine learning algorithms. However, these models often struggle to generalize across different contexts, necessitating repeated cycles of data collection, annotation, and training. The emergence of Vision-Language Models (VLMs) offers a promising alternative, enabling generalization across visual recognition tasks through zero-shot prompting without requiring fine-tuning. This study investigates the potential of VLMs to analyze students' academic emotions via facial expressions in an online learning environment. We employed two VLMs, Llama-3.2-11B-Vision-Instruct and Qwen2.5-VL-7B-Instruct, to analyze 5,000 images depicting confused, distracted, happy, neutral, and tired expressions using zero-shot prompting. Preliminary results indicate that both models demonstrate moderate performance in academic facial expression recognition, with Qwen2.5-VL-7B-Instruct outperforming Llama-3.2-11B-Vision-Instruct. Notably, both models excel in identifying students' happy emotions but fail to detect distracted behavior. Additionally, Qwen2.5-VL-7B-Instruct exhibits relatively high performance in recognizing students' confused expressions, highlighting its potential for practical applications in identifying content that causes student confusion.
\end{abstract}


\section{Introduction}
Academic emotions refer to the emotions and feelings experienced by students in educational settings, such as enjoyment of learning, pride in success, and test anxiety \cite{pekrun2002academic}. These emotions have been found to significantly impact students' learning outcomes \cite{alshareef2024role,ekornes2022impact,tan2021influence}. For instance, a meta-analysis indicates a significantly positive correlation between positive emotions and academic achievement, and a significantly negative correlation between negative emotions and academic achievement \cite{lei2016effects}. Moreover, extreme negative academic emotions, such as depression resulting from academic stress, can lead to self-harming behaviors and, in the most severe cases, suicide \cite{guo2019association}.

Given the significance of academic emotions, researchers have explored various methods to assess these emotions in students. Traditionally, questionnaires have been employed as self-report tools for students to disclose their academic emotions \cite{chiang2014scale,pekrun2014self}. However, due to fear of negative consequences, such as criticism from teachers and parents, students may sometimes conceal their true emotions and report their feelings in a more positive light. Additionally, static questionnaires are limited in their ability to capture the dynamic fluctuations of emotions during academic activities \cite{kubsch2022once}.

To address these limitations, researchers have begun to evaluate students' academic emotions through observation using multimodal approaches \cite{abdullah2021multimodal,dubovi2022cognitive,zhu2024review}. For instance, physiological signals have been employed as indicators to reflect students' emotional states and their variations during learning \cite{ketonen2023can,wei2021application}. However, the high cost of biological sensors and the impracticality of requiring students to wear these sensors consistently pose significant challenges for schools. In addition to physiological signals, other modalities such as speech, text, images, and videos have been collected to analyze students' academic emotions through voice, utterances, facial expressions, gestures, and postures \cite{lek2023academic,zhu2024review}.

Faced with large volumes of data, researchers initially adopted traditional machine learning algorithms, such as support vector machines, decision trees, and random forests, to classify students' academic emotions \cite{lek2023academic,rajan2019facial}. Although these methods provided automated analysis, their performance required further improvement. With advancements in deep learning, deep neural networks, including convolutional neural networks (CNNs), long short-term memory networks (LSTMs), and Transformers, have been utilized to detect students' text sentiment, facial expressions, body gestures, and postures, thereby enhancing the analysis of academic emotions \cite{bian2019spontaneous,feng2020academic,lek2023academic,xiang2024multimodal}. Despite the outstanding performance of deep learning models, they may not generalize well across different contexts. For example, a model trained to recognize facial expressions may not be effective for detecting gestures and postures. Furthermore, models trained on Asian students may not perform adequately when applied to students from Western countries due to their differences in facial features.

With the advancement of artificial intelligence (AI), the emergence of vision-language models (VLMs) presents a promising solution to the aforementioned issues. Pre-trained on large-scale datasets of image-text pairs, VLMs are multimodal models capable of performing various visual recognition tasks via zero-shot prompting without the need for fine-tuning \cite{zhang2024vision}. For example, VLMs can take images and text as inputs and accordingly generate text-based answers to user questions \cite{bordes2024introduction}. However, the application of VLMs in detecting students' academic emotions remains largely unexplored. Thus, this study investigates the potential of two VLMs, Llama-3.2-11B-Vision-Instruct and Qwen2.5-VL-7B-Instruct, in detecting students' academic emotions by analyzing their facial expressions. 

\begin{figure*}[h]
    \centering
    \includegraphics[width=0.8\linewidth]{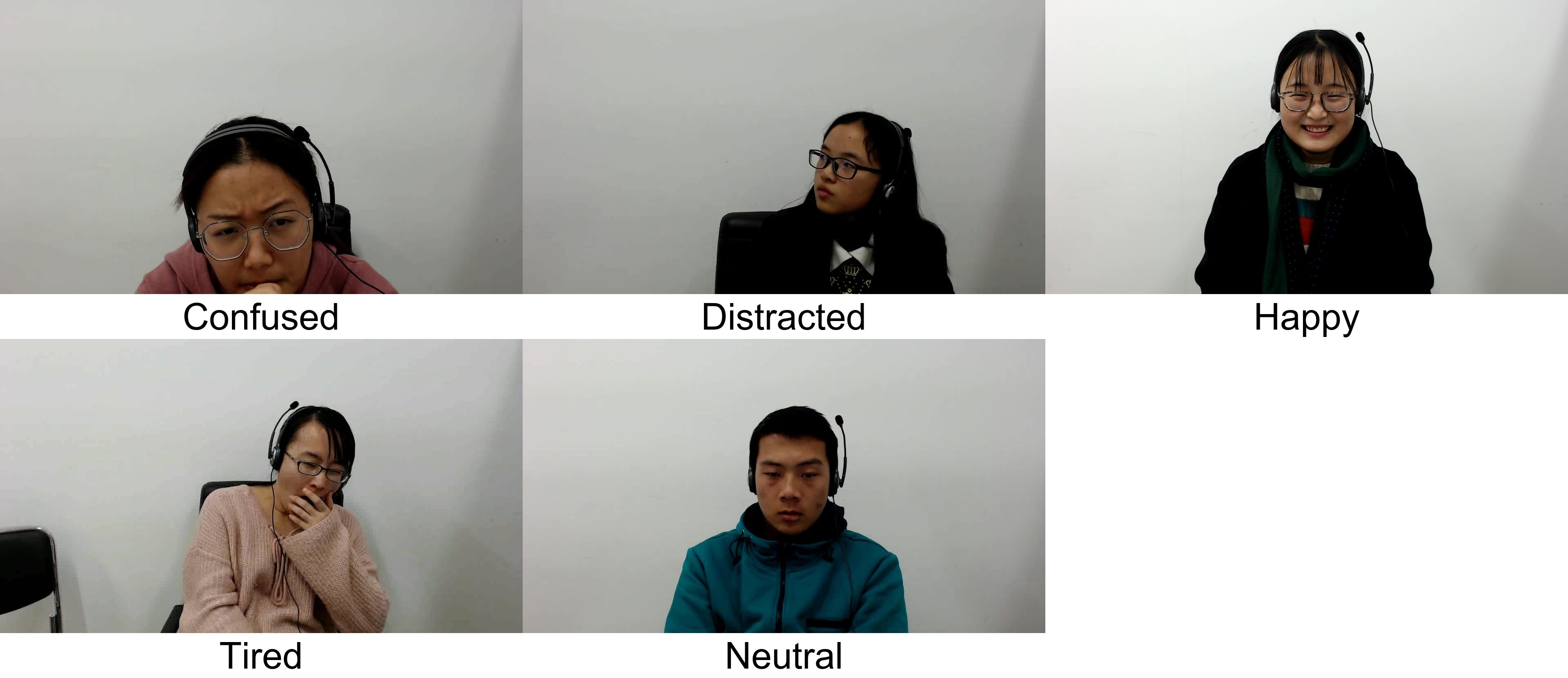}
    \caption{Five facial expressions in the OLSFED dataset\cite{bian2019spontaneous}.}
    \label{fig:examples}
\end{figure*}
\begin{figure*}[h]
    \centering
    \includegraphics[width=0.89\linewidth]{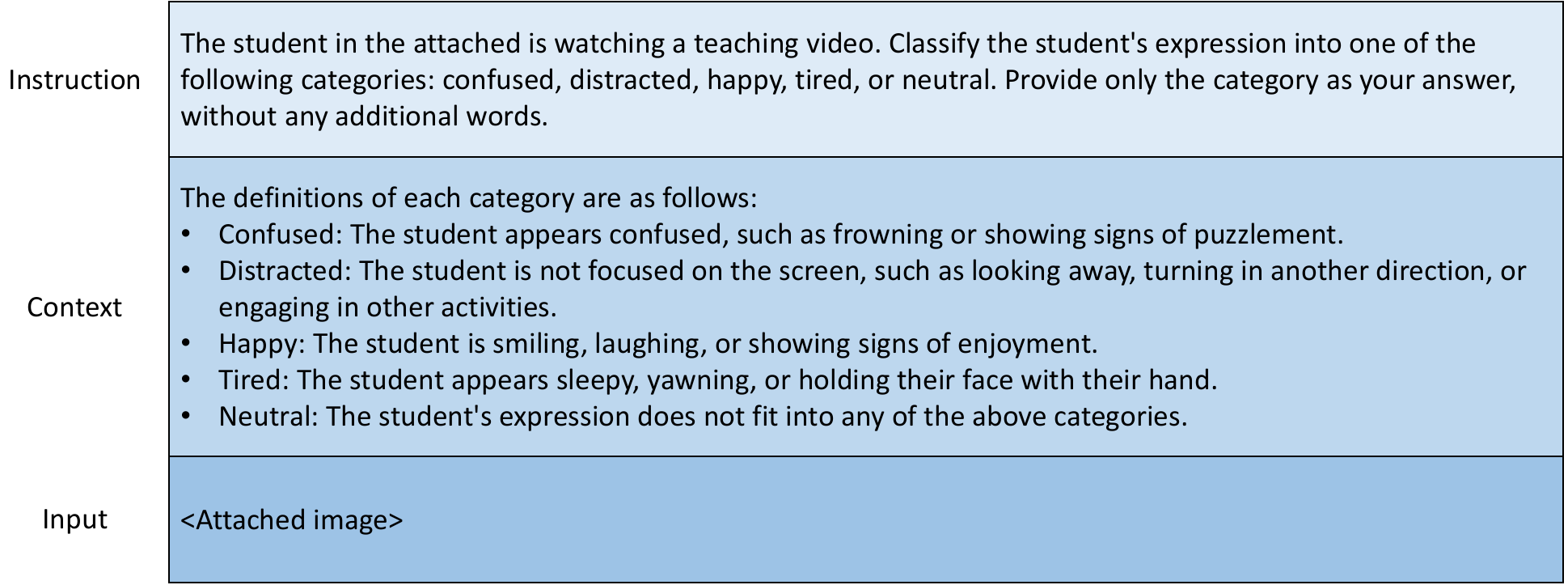}
    \caption{The prompt designed for VLMs.}
    \label{fig:prompt}
\end{figure*}

\section{Related Work}

In 1998, the American Educational Research Association held an annual meeting on ``The Role of Emotions in Students’ Learning and Achievement," which marked the beginning of significant attention being paid to students' emotions by researchers. Numerous studies have since found that students experience a variety of emotions in academic settings and have further confirmed that these academic emotions can impact students’ learning outcomes \cite{alshareef2024role,ekornes2022impact,pekrun2016academic,tan2021influence}. Moreover, academic emotions have also been found to influence students' social behavior \cite{guo2019association}. Therefore, it is crucial to accurately identify students' emotions in academic scenarios to facilitate appropriate interventions.

Traditionally, questionnaires have been the predominant method for measuring students' academic emotions, such as the achievement emotions questionnaire \cite{pekrun2011measuring} and the academic emotion scale \cite{govaerts2008development}. However, given that questionnaires cannot capture the dynamic changes in students' academic emotions, researchers have begun to employ technologies to dynamically track students' academic emotions in a multimodal manner. For instance, psychophysiological measures, such as electrodermal activity and heart rate, have been used as indicators to identify and monitor students' academic emotions in real-time during learning in computer-based learning environments \cite{huber2023happens}.

In addition, students' academic emotions can be analyzed through speech, text, images, and videos. For example, continuous speech within classroom settings has been utilized to build a speech emotion recognition system for online learning \cite{cen2016real}. Transcripts of students' speech can be used to trace the evolution of students' sentiments in classroom learning \cite{huang2021sentiment} and computer-supported collaborative learning \cite{zheng2023automated}. Images of students' performance during learning can be used to recognize their academic emotions through facial expressions, gestures, and postures \cite{castellano2008emotion,lek2023academic}. Furthermore, videos of classroom teaching can be employed to analyze students' emotional engagement in learning \cite{zeng2020emotioncues}.

To automate the analysis of academic emotions, researchers commonly use supervised machine learning techniques, including traditional algorithms and deep learning algorithms, to train models. However, these models have several limitations. For example, models trained to recognize facial expressions cannot be used to analyze body gestures. Additionally, models trained on Asian students may not perform well on Western students due to differing facial traits. In such cases, researchers may need to repeat the data collection, data annotation, and model training processes. The emergence of VLMs appears to offer a solution to these issues due to their ability to generalize across visual recognition tasks via zero-shot prompting. Therefore, this study aims to explore whether VLMs can be used to identify students' academic emotions through facial expressions.

\section{Method}
\subsection{Dataset}
In this study, we analyzed students' academic emotions through their facial expressions. Specifically, we utilized the Online Learning Spontaneous Facial Expression Database (OLSFED) \cite{bian2019spontaneous} as our academic emotion dataset. This dataset captures the facial expressions of 82 students while they watched instructional videos in an online learning context, comprising a total of 31,115 images representing five facial expressions: confused, distracted, happy, tired, and neutral, as illustrated in Figure \ref{fig:examples}. Each image has a resolution of 1280x720, ensuring high quality. All participants were Asian students, signed consent forms permitting the use of their images and videos for non-commercial scientific research. For this preliminary study, we randomly selected 1,000 images for each type of facial expression for analysis by the VLMs, resulting in a total of 5,000 images.


\subsection{Prompting vision language models}
Llama-3.2-11B-Vision-Instruct\footnote{https://huggingface.co/meta-llama/Llama-3.2-11B-Vision-Instruct} and Qwen2.5-VL-7B-Instruct\footnote{https://huggingface.co/Qwen/Qwen2.5-VL-7B-Instruct} were selected as the representative VLMs in this study due to their open-source nature. Specifically, Llama-3.2-11B-Vision-Instruct, released in September 2024, is optimized from the pre-trained Llama 3.1 language model for tasks including visual recognition, image reasoning, captioning, and answering general questions about images \cite{Llama-3.2-11B-Vision-Instruct}. This model outperforms many available open-source and proprietary multimodal models on standard industry benchmarks. Similarly, Qwen2.5-VL-7B-Instruct, released in January 2025, is optimized from Qwen2-VL for tasks such as recognizing and locating objects within images, analyzing image content, and comprehending long videos and pinpointing specific video segments \cite{qwen2.5-VL}. This model also achieves outstanding performance on standard industry benchmarks.


The prompt designed for Llama-3.2-11B-Vision-Instruct and Qwen2.5-VL-7B-Instruct to analyze students' facial expressions comprises three components: \textit{instruction}, \textit{context}, and \textit{input}, as illustrated in Figure \ref{fig:prompt}. The \textit{instruction} component specifies the task for the VLMs. In this study, the instruction was formulated as follows: \textit{The student in the attached image is watching a teaching video. Classify the student's expression into one of the following categories: confused, distracted, happy, tired, or neutral.} To facilitate the swift extraction of the classification result, an additional directive was included: \textit{Provide only the category as your answer, without any additional words.} This precaution was necessary to prevent VLMs from generating lengthy responses, which could complicate the accurate and prompt extraction of the classification results. The \textit{context} component provides definitions for the five facial expressions to enhance classification accuracy. Finally, the \textit{input} component consists of the student's images to be analyzed.

\section{Results}
\begin{table*}[]
\caption{The overall performance of Llama-3.2-11B-Vision-Instruct and Qwen2.5-VL-7B-Instruct.}
\label{tab:overall}
\centering
\begin{tabular}{lllll}
\hline
Model                         & F1 Score & Accuracy & Precision & Recall \\ \hline
Llama-3.2-11B-Vision-Instruct & 0.4519   & 0.5118   & 0.6473    & 0.5118 \\
Qwen2.5-VL-7B-Instruct        & 0.5197   & 0.5546   & 0.7613    & 0.5546 \\ \hline
\end{tabular}
\end{table*}
\begin{figure}[h]
    \centering
    \includegraphics[width=0.9\linewidth]{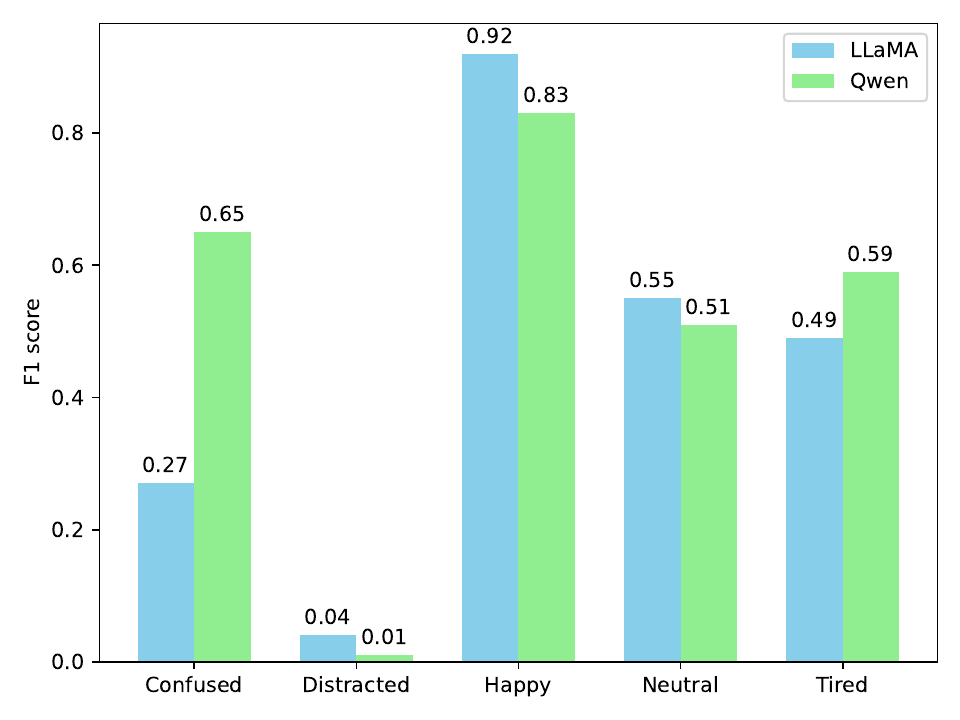}
    \caption{F1 scores of of Llama-3.2-11B-Vision-Instruct and Qwen2.5-VL-7B-Instruct on four facial expressions.}
    \label{fig:f1score}
\end{figure}
\begin{figure}[h]
    \centering
    \includegraphics[width=0.9\linewidth]{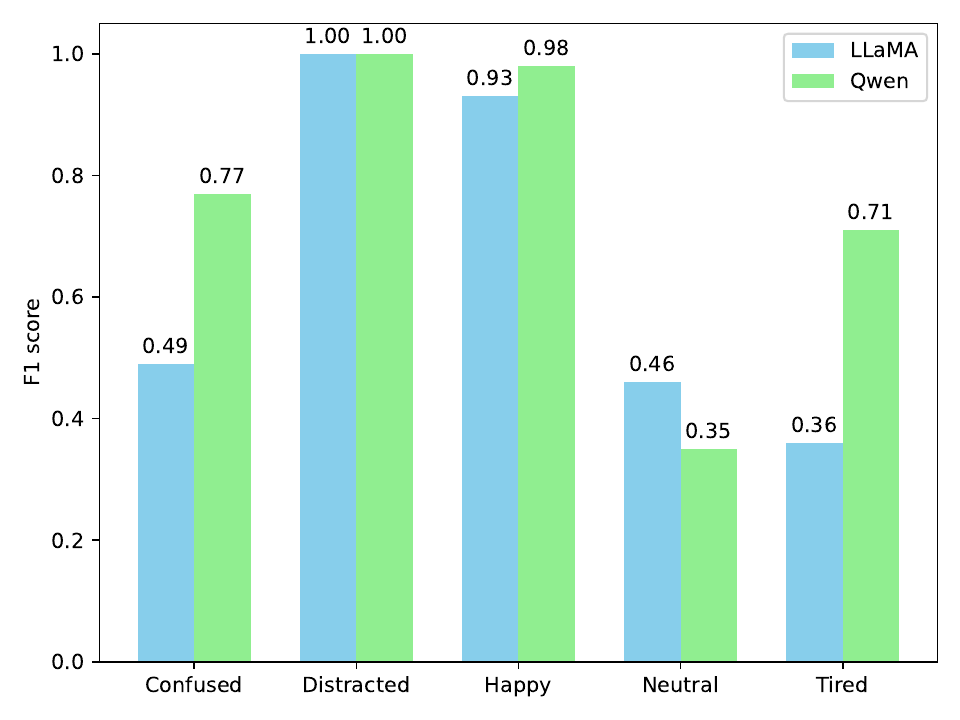}
    \caption{Precision scores of Llama-3.2-11B-Vision-Instruct and Qwen2.5-VL-7B-Instruct on four facial expressions.} 
    \label{fig:precision} 
\end{figure}
\begin{figure}[h]
    \centering
    \includegraphics[width=0.9\linewidth]{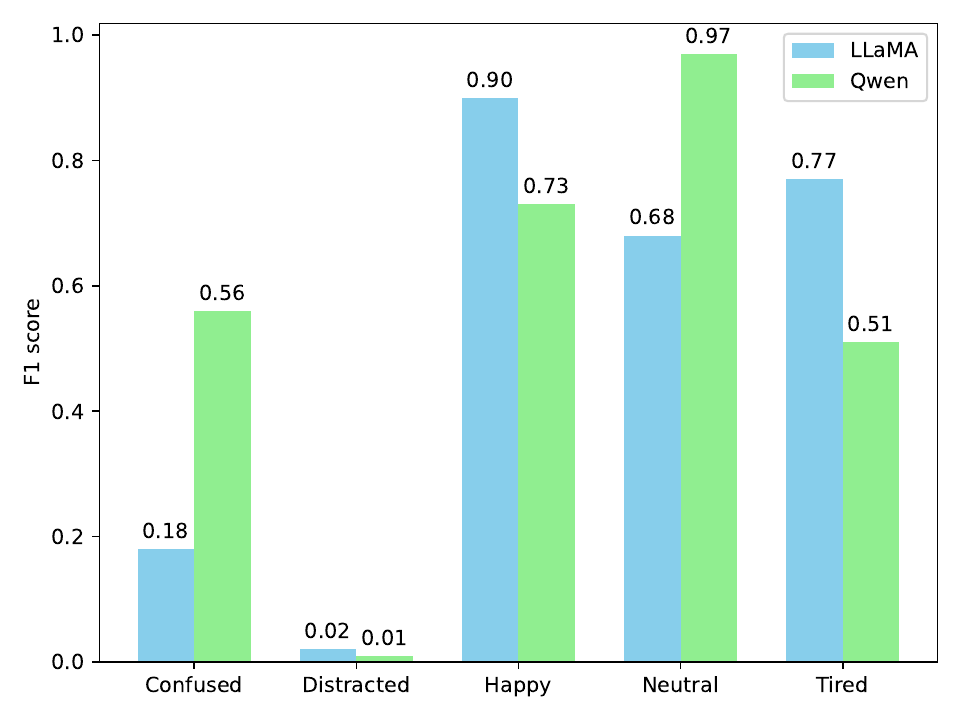}
    \caption{Recall scores of Llama-3.2-11B-Vision-Instruct and Qwen2.5-VL-7B-Instruct on four facial expressions.}
    \label{fig:recall}
\end{figure}

Table \ref{tab:overall} presents the overall performance of Llama-3.2-11B-Vision-Instruct and Qwen2.5-VL-7B-Instruct in classifying five academic facial expressions. Specifically, Llama-3.2-11B-Vision-Instruct achieves 0.4519 in F1 score, 0.5118 in accuracy, 0.6473 in precision, and 0.5118 in recall, respectively. In contrast, Qwen2.5-VL-7B-Instruct achieves corresponding values of 0.5197, 0.5546, 0.7613, and 0.5546. Given that facial expression recognition in our study constitutes a five-way classification task, we consider the F1 score and accuracy as the primary metrics for evaluating model performance. Consequently, both models demonstrate moderate performance, with Qwen2.5-VL-7B-Instruct outperforming Llama-3.2-11B-Vision-Instruct.

Figures \ref{fig:f1score}, \ref{fig:precision}, and \ref{fig:recall} illustrate the F1 score, precision, and recall for each facial expression category for both models. Specifically, for the F1 score, Llama-3.2-11B-Vision-Instruct achieves values of 0.27, 0.04, 0.92, 0.55, and 0.49 across the five facial expression categories, whereas Qwen2.5-VL-7B-Instruct achieves values of 0.65, 0.01, 0.83, 0.51, and 0.59, respectively. Both models perform well in recognizing the 'happy' emotion but fail to detect the 'distracted' behavior. They perform moderately in identifying 'neutral' and 'tired' expressions. Moreover, Llama-3.2-11B-Vision-Instruct exhibits significant limitations in recognizing the 'confused' expression, while Qwen2.5-VL-7B-Instruct demonstrates relatively high performance in this category.

Similar trends are observed in the precision and recall metrics. Notably, both models achieve high precision scores (1.00 and 1.00) for the 'distracted' expression but exhibit extremely low recall scores (0.02 and 0.01) for the same category. This indicates that although the models classify few images as 'distracted,' these classifications are accurate.

\section{Discussion and Conclusion}

To automatically and accurately analyze students' academic emotions, researchers have utilized supervised machine learning algorithms to train various models. However, these models face challenges in generalizing well across different contexts, necessitating repeated cycles of data collection, annotation, and training. The advent of VLMs offers a potential solution to these issues, as they can generalize across visual recognition tasks via zero-shot prompting without the need for fine-tuning. Consequently, this study explores the potential of VLMs to analyze students' academic emotions through facial expressions in an online learning environment. Specifically, we employed Llama-3.2-11B-Vision-Instruct and Qwen2.5-VL-7B-Instruct to analyze 5,000 images representing confused, distracted, happy, neutral, and tired expressions using zero-shot prompting. The preliminary results indicate that both models demonstrate moderate performance in the task of academic facial expression recognition, with Qwen2.5-VL-7B-Instruct outperforming Llama-3.2-11B-Vision-Instruct. Notably, both models excel in identifying students' happy emotions but fail to effectively detect distracted behavior. Moreover, Qwen2.5-VL-7B-Instruct shows relatively high performance in identifying students' confused expressions, suggesting its significant potential for practical applications in identifying content that causes student confusion.

Despite these promising findings, several limitations of this exploratory study must be acknowledged. First, the preliminary  analysis was conducted on a subset of an academic emotion dataset. The findings would be more robust with experiments conducted on more comprehensive datasets. Second, this study focused solely on analyzing academic emotions through facial expressions. In reality, academic emotions can be analyzed through multiple modalities. Therefore, the potential of VLMs for academic emotion analysis could also be explored through other modalities, such as students' gestures. Third, only two VLMs were examined in this study. Future research should investigate a broader range of VLMs, including proprietary models. Consequently, promising research directions in this area include leveraging multiple VLMs to analyze academic emotions through multimodal data on diverse datasets and evaluating their practical impact.
\bibliographystyle{abbrv}
\bibliography{sigproc}

\end{document}